\documentclass[10pt,twocolumn,letterpaper]{article}

\usepackage{iccv}
\usepackage{times}
\usepackage{epsfig}
\usepackage{graphicx}
\usepackage{amsmath}
\usepackage{amssymb}
\usepackage{algorithm}
\usepackage{algorithmic}
\usepackage{bbm}
\usepackage{amsfonts}
\usepackage{mathrsfs}
\usepackage{multirow}

\usepackage{booktabs}
\usepackage{enumitem}

\usepackage[pagebackref=true,breaklinks=true,letterpaper=true,colorlinks,bookmarks=false]{hyperref}

\iccvfinalcopy 

\def\iccvPaperID{406} 

\title{Scalable Autoregressive Monocular Depth Estimation}

\author{Jinhong Wang\textsuperscript{\rm 1} \quad Jian Liu\textsuperscript{\rm 2} \quad Dongqi Tang\textsuperscript{\rm 2} \quad Weiqiang Wang\textsuperscript{\rm 2} \quad Wentong Li\textsuperscript{\rm 1} \\ Danny Chen\textsuperscript{\rm 3} \quad Jintai Chen\textsuperscript{\rm 4$\dagger$} \quad Jian Wu\textsuperscript{\rm 1$\dagger$} \vspace{0.3cm} \\
    \textsuperscript{\rm 1}ZJU
        \quad \quad \textsuperscript{\rm 2}Ant Group
    \quad \quad \textsuperscript{\rm 3}University of Notre Dame
    \quad \quad \textsuperscript{\rm 4}HKUST(Guangzhou) \\  
    $\dagger$ {\small Corresponding Authors}
}

\begin{document}


\maketitle

\begin{abstract}
This paper shows that the autoregressive model is an effective and scalable monocular depth estimator. Our idea is simple: We tackle the monocular depth estimation (MDE) task with an autoregressive prediction paradigm, based on two core designs. First, our depth autoregressive model (DAR) treats the depth map of different resolutions as a set of tokens, and conducts the \texttt{low-to-high} resolution autoregressive objective with a patch-wise causal mask. Second, our DAR recursively discretizes the entire depth range into more compact intervals, and attains the \texttt{coarse-to-fine} granularity autoregressive objective in an ordinal-regression manner. By coupling these two autoregressive objectives, our DAR establishes new state-of-the-art (SOTA) on \texttt{KITTI} and \texttt{NYU Depth v2} by clear margins. Further, our scalable approach allows us to scale the model up to 2.0B and achieve the best RMSE of 1.799 on the \texttt{KITTI} dataset (5\% improvement) compared to 1.896 by the current SOTA (\textit{Depth Anything}). DAR further showcases zero-shot generalization ability on unseen datasets. These results suggest that DAR yields superior performance with an autoregressive prediction paradigm, providing a promising approach to equip modern autoregressive large models (e.g., GPT-4o) with depth estimation capabilities.
\end{abstract}

\section{Introduction}
\label{sec:intro}

\begin{figure}[t]
\centering
\includegraphics[width=0.43\textwidth]{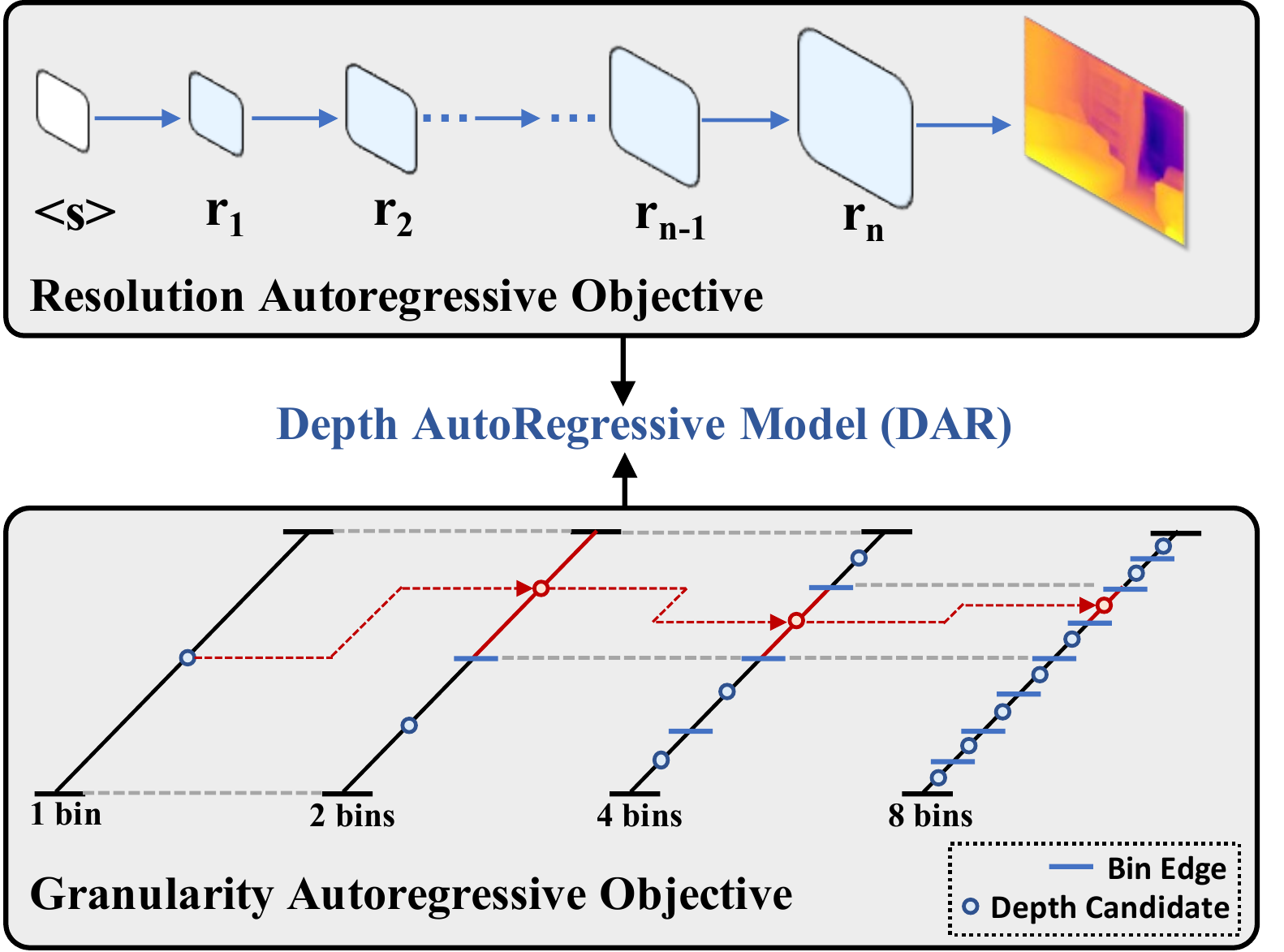}
\caption{We exploit two ``order" properties of the MDE task that can be transformed into two autoregressive objectives. {\bf (a) Resolution autoregressive objective:} The generation of depth maps can follow a resolution order from low to high. For each step of the resolution autoregressive process, the Transformer predicts the next higher-resolution token map conditioned on all the previous ones. {\bf (b) Granularity autoregressive objective:} The range of depth values is ordered, from 0 to specific max values. For each step of the granularity autoregressive process, we increase exponentially the number of bins (e.g., doubling the bin number), and utilize the previous predictions to predict a more refined depth with a smaller and more refined granularity. Our proposed DAR aims to perform these two autoregressive processes simultaneously.}
\label{fig1}
\vskip -1 em
\end{figure}

The Monocular Depth Estimation (MDE) task aims to predict per pixel depth from a single RGB image, which plays a crucial role in scene understanding and reconstruction~\cite{izadi2011kinectfusion,chen2019towards}. This task has broad applications, including autonomous driving~\cite{wang2019pseudo}, robotics~\cite{jia2023object}, augmented reality~\cite{oney2020evaluation}, medical endoscopic surgery~\cite{liu2019dense}, etc.
Most deep learning (DL) based methods typically followed top-down-bottom-up encoder-decoder architectures~\cite{eigen2014depth,laina2016deeper,yuan2022neural,patil2022p3depth,liu2023va,aich2021bidirectional,yang2021transformer,agarwal2023attention,li2023depthformer,duan2023diffusiondepth,ji2023ddp,zhao2023unleashing,saxena2023monocular,ke2024repurposing}, which extract and fuse low-level and high-level features for depth estimation.

Recently, autoregressive (AR) architectures have demonstrated strong generalization capabilities and significant scalability across a multitude of tasks: (i) Generalization: AR models achieve remarkable zero- and few-shot performance on previously unseen datasets, and exhibit substantial flexibility in adapting to diverse downstream tasks~\cite{radford2019language,brown2020language}. (ii) Scalability: As suggested by scaling laws~\cite{kaplan2020scaling,henighan2020scaling}, AR models allow flexible model size scaling for optimal performance across various practical applications. Recent advances in large language models (LLMs)~\cite{radford2018improving,radford2019language,brown2020language} and multi-modal large language models (MLLMs) (e.g., GPT-4 and LLaVA~\cite{achiam2023gpt,liu2024visual}), built upon AR architectures, achieved outstanding results in a wide range of tasks, including text-to-image generation~\cite{esser2021taming,lee2022autoregressive,tian2024visual}, video generation~\cite{yan2021videogpt,wu2022nuwa}, object detection~\cite{chen2021pix2seq} and tracking~\cite{wei2023autoregressive}, among others.

This naturally leads to an interesting question: \textit{Can an autoregressive model be developed for 
the MDE task?}

However, autoregressive modeling relies on a well-organized sequential data formation, where each step’s prediction is logically connected to the previous steps. While this sequential dependency may be common in other tasks, it is not intuitively align with the  MDE requirements, where meaningful sequential prediction targets are less apparent.


\begin{figure}[t]
\centering
\includegraphics[width=0.47\textwidth]{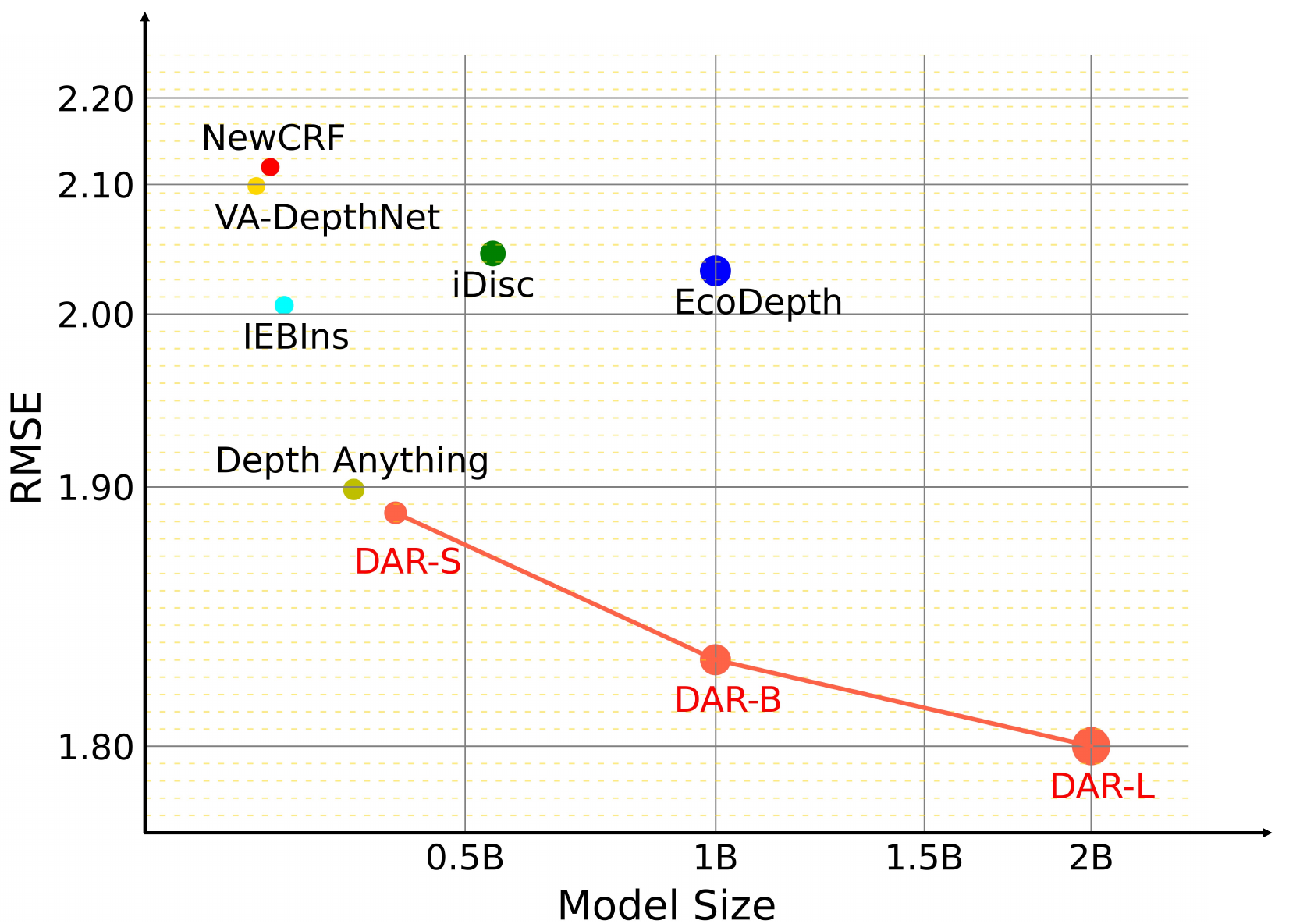}
\caption{RMSE performances ($\downarrow$) {\textit vs.} model sizes on the \texttt{KITTI} dataset. Our DAR shows strong scalability and achieves better performance-efficiency trade-off among cutting-edge methods.}
\label{figp}
\vskip -1 em
\end{figure}

In this paper, we introduce a simple, effective, and scalable depth autoregressive (DAR) framework for MDE. Our new approach leverages two key ordering properties of MDE, and we incorporate them as autoregressive objectives, as shown in Fig.~\ref{fig1}. The first property is ``depth map resolution'': We generate depth maps at varying resolutions, ordered from low to high, and treat the sequences of depth maps of different resolutions as prediction targets~\cite{tian2024visual}. This approach reframes depth map generation as a \texttt{low-to-high} resolution autoregressive objective, where each step generates a higher-resolution depth map based on previous predictions. The second property is ``depth values", which inherently exist in a continuous space. Typically, known methods discretize the depth values into several intervals (or bins) to formulate an ordinal regression task~\cite{fu2018deep}. By further discretizing the depth range into progressively finer intervals, we recast MDE as a \texttt{coarse-to-fine} autoregressive objective. 

For the resolution autoregressive objective, we develop a depth autoregressive Transformer that predicts the next-resolution depth map based on its prefix predictions with the patch-wise causal mask. To achieve the granularity autoregressive objective, we propose a novel binning strategy called Multiway Tree Bins (MTBin), which queries the corresponding bin using prior depth predictions and then recursively refines each bin into sub-bins with error tolerance for subsequent autoregressive steps. Importantly, these bins are used to not only compute the final depth values but also embed granularity information into the latent token maps, effectively guiding the depth map generation process.

Comprehensive experiments show that our DAR achieves state-of-the-art (SOTA) performance on the \texttt{KITTI} and \texttt{NYU Depth v2} datasets. With a similar model size, DAR outperforms current SOTA (\textit{Depth Anything}) in all the metrics, particularly by 3\% in the RMSE metric on \texttt{KITTI}. DAR also exhibits scaling laws akin to those witnessed in LLMs, and its size can be easily scaled up to 2.0B, which establishes a new SOTA performance, as shown in Fig.~\ref{figp}. Lastly, we showcase DAR’s zero-shot generalization capabilities on unseen datasets. These results further validate the generalizability and scalability of DAR.

\begin{figure*}[t]
\centering
\includegraphics[width=0.85\textwidth]{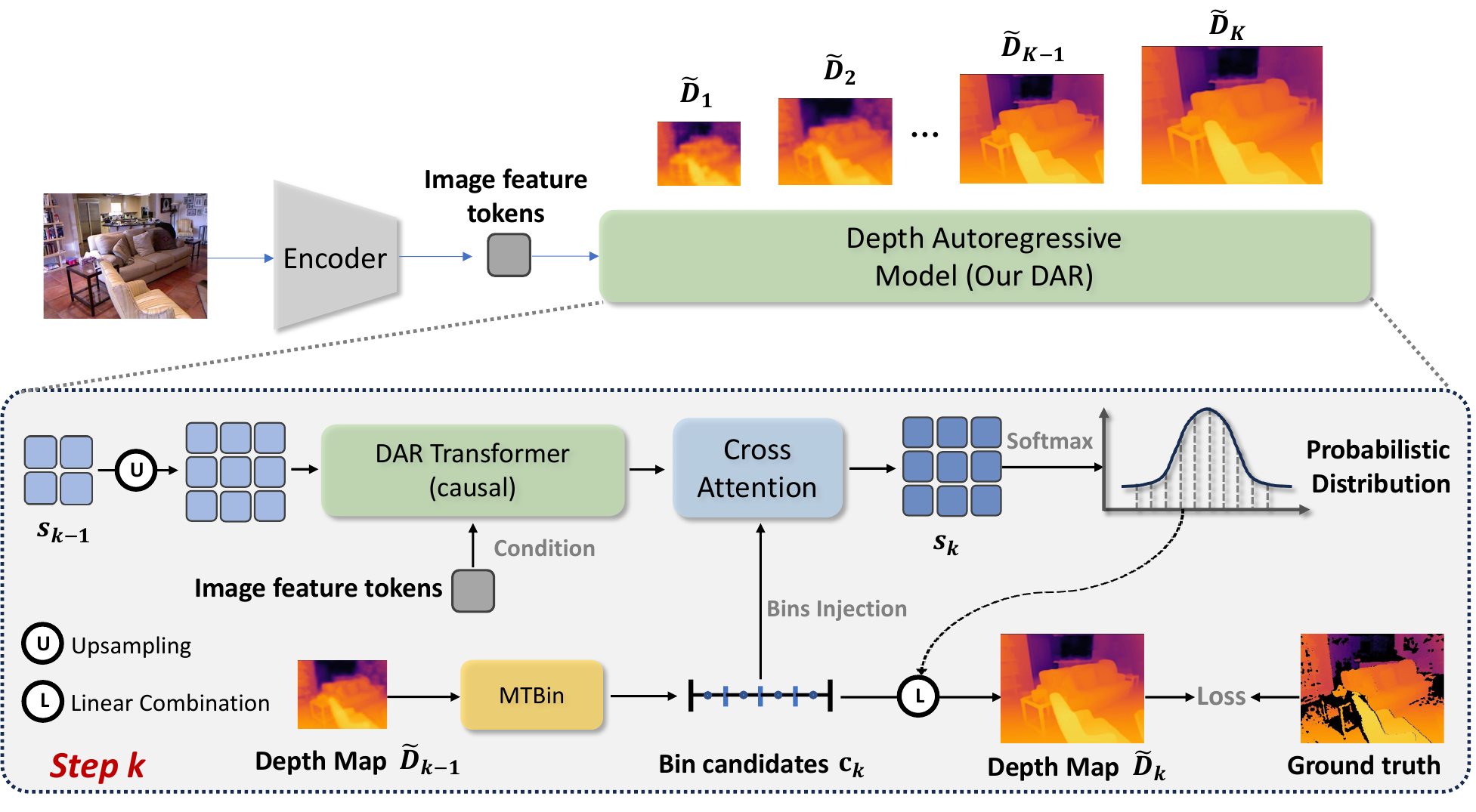}
\caption{{\bf An overview of DAR.} We begin with encoding the input RGB images into image tokens as the context condition. At each step, DAR Transformer with the patch-wise causal mask performs autoregressive predictions, that is, it allows the input token map (upsampled from the previous resolution token map $r_{k-1}$) to interact with only the prefix tokens and global image feature tokens for the next-resolution token map modeling. The output latent tokens are then sent to the ConvGRU module, which injects the prompts of new refined bin candidates $\mathbf{c}_{k}$ (generated by MTBin from the previous prediction $\tilde{D}_{k-1}$) for further granularity guidance and generates the next-resolution token map $r_{k}$. The new depth map $\tilde{D}_{k}$ is generated by a linear combination of the next-granularity bin candidates $\mathbf{c}_{k}$ and softmax value $\mathbf{p}_{k}$ of the next-resolution token map, achieving concurrently a resolution and granularity autoregressive evolution. }
\label{fig2}
\vskip -1 em
\end{figure*}

The contributions of this work are multi-fold:
\noindent
\begin{itemize}
\item[{\bf (1)}] To the best of our knowledge, we introduce the first autoregressive model for monocular depth estimation (MDE), called DAR — a simple, effective, and scalable framework. Our key insight lies in transforming two ordered properties in MDE, depth map resolution and granularity, into autoregressive objectives.

\item[{\bf (2)}] We reformulate the low- and high-level feature fusion process in existing encoder-decoder models as a \texttt{low-to-high} resolution autoregressive objective. We introduce a new Depth autoregressive Transformer that uses the patch-wise causal mask and progressively generates depth maps at increasing resolutions, conditioned on tokens from the input RGB images.

\item[{\bf (3)}] We propose a novel binning strategy called multiway tree bins (MTBin), tailored for the granularity autoregressive objective, which transforms MDE tasks into an autoregressive bin sequence prediction task. By embedding bin information into the latent token map, DAR effectively connects the resolution and granularity autoregressive processes.

\item[{\bf (4)}] DAR establishes new state-of-the-art performance on the \texttt{KITTI} and \texttt{NYU Depth v2} datasets, and exhibits stronger zero-shot capability than \texttt{Depth Anything}~\cite{yang2024depth}. A series of DAR models, ranging from 440M to 2.0B parameters, is developed based on decoder-only Transformer~\cite{vaswani2017attention,radford2019language}. The largest model achieves 0.205 RMSE and 0.982 $\delta_{1}$ on \texttt{NYU Depth v2}, largely outperforming the existing methods.
\end{itemize}

\section{Related Work}
\label{sec:Rel}

\noindent
{\bf Monocular Depth Estimation (MDE).}
Monocular Depth Estimation has a long development of methods, ranging from traditional methods to deep learning (DL) techniques. Traditional methods relied on handcrafted features~\cite{hoiem2007recovering,liu2008sift} and used Markov Random Fields~\cite{saxena2007learning} to predict depth maps.  However, they suffered limitations in handling complex scenes.
Modern DL techniques approached the problem as a dense regression problem. Successive improvements mainly came from three fronts: model architecture, data-driven, and language guidance. {\bf (1) Model architecture:} The main improvement came from the transformation of model backbones, from CNNs~\cite{eigen2014depth,laina2016deeper,yuan2022neural,patil2022p3depth,liu2023va} to Transformers~\cite{aich2021bidirectional,yang2021transformer,agarwal2023attention,li2023depthformer} to the current diffusion models~\cite{duan2023diffusiondepth,ji2023ddp,zhao2023unleashing,saxena2023monocular,ke2024repurposing}. {\bf (2) Data-driven methods:} The milestone data-driven method MiDas~\cite{ranftl2020towards} proposed to train well-generalized models on large amounts of data by mixing different datasets, which achieved excellent zero-shot transfer performance. ZoeDepth~\cite{bhat2023zoedepth} and Depth Anything~\cite{yang2024depth} further leveraged the benefits of large-scale unlabeled data (62M) via self-supervised learning to obtain outstanding zero-shot generalization.  {\bf (3) Language-guided methods:} Benefited from the rich visual and text information of CLIP and other language-vision pre-training models, VPD and other methods~\cite{zhao2023unleashing,zeng2024wordepth,patni2024ecodepth,chatterjee2024robustness} used language description to facilitate depth estimation and achieved SOTA results on standard depth estimation datasets.

\noindent
{\bf MDE as Ordinal Regression.}
Another main line of work is to treat MDE as an ordinal regression task~\cite{herbrich1999support,harrell2012regression}, which aims to predict labels on an ordinal scale. By discretizing the depth space into several bins, DORN~\cite{fu2018deep} first proposed an ordinal regression network to predict discrete depth values. Since then, many studies~\cite{bhat2021adabins,shao2024iebins,li2024binsformer} developed various depth binning strategies to discretize depth range and solve MDE as ordinal regression. In particular, Ord2Seq~\cite{wang2023ord2seq} treated ordinal regression as a label sequence task, and proposed an autoregressive network to predict the more refined labels progressively. Inspired by this, we transform MDE as an autoregressive prediction task from a bin perspective to achieve more granular prediction progressively.

\noindent
{\bf Autoregressive Visual Generation.}
Many recent methods~\cite{esser2021taming,razavi2019generating,lee2022autoregressive} explored the effectiveness of autoregressive models in the visual domain. VQGAN~\cite{esser2021taming} proposed to use VQVAE~\cite{van2017neural} to conduct the autoregressive process in the latent space. It employed GPT-2 decoder-only Transformer to generate tokens in the raster-scan order. VQVAE-2~\cite{razavi2019generating} and RQ-Transformer~\cite{lee2022autoregressive} also followed this raster-scan manner, but used extra scales or stacked codes. VAR~\cite{tian2024visual} proposed text-to-image autoregressive generation via next-scale prediction, which transforms the entire image into a set of tokens and treats them as input to predict the next-scale target image. Inspired by this, we transform MDE as an autoregressive prediction task from a scale perspective to achieve larger resolution predictions progressively.

\begin{figure}[t]
\centering
\includegraphics[width=0.45\textwidth]{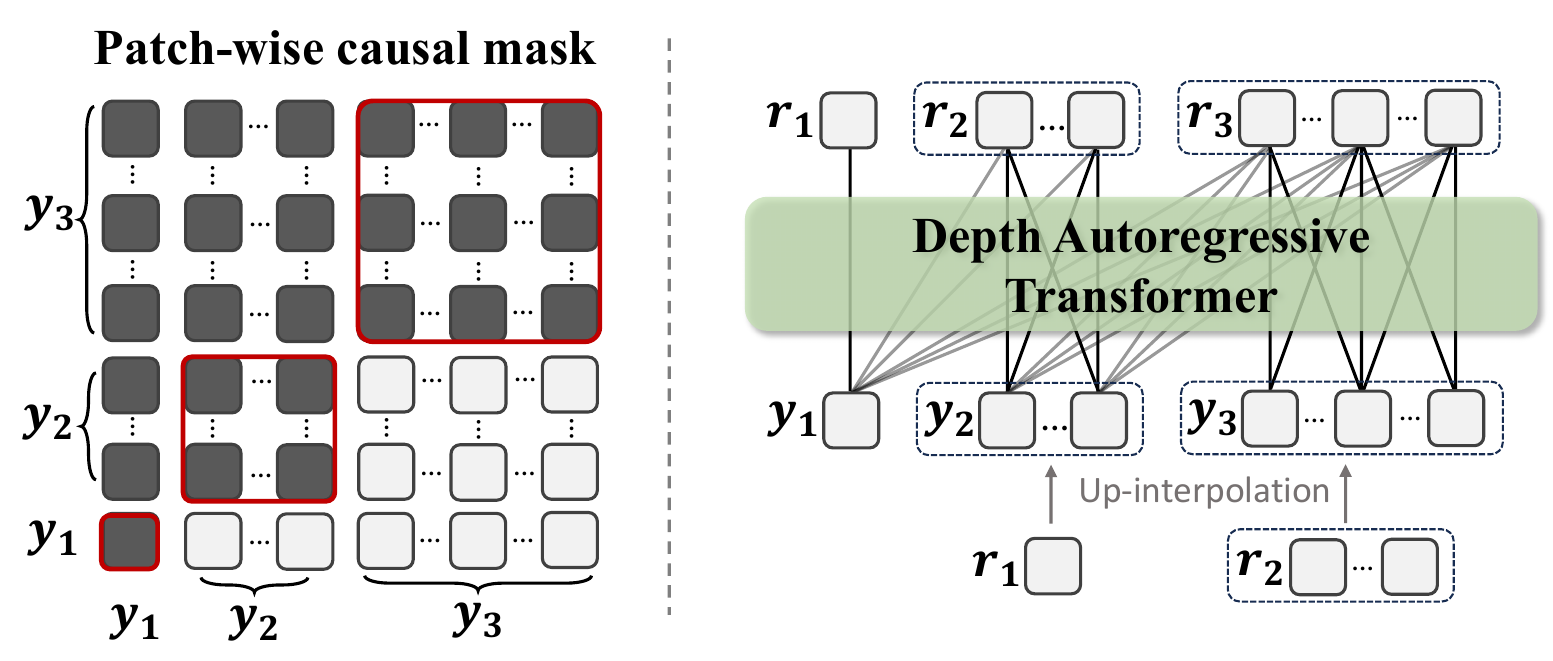}
\caption{Illustrating the patch-wise causal mask for ensuring that the current token map can interact only with tokens from itself and the prefix token maps.}
\label{fig4}
\vskip -1 em
\end{figure}

\section{Method}
\label{sec:method}

\subsection{Preliminaries}

\noindent
{\bf Ordinal Regression-based MDE.}
Some methods~\cite{fu2018deep,bhat2021adabins} tackled MDE in an ordinal regression fashion, learning the probabilistic distribution on each pixel and using a linear combination with depth candidates as the final depth prediction. Suppose we divide the depth range $[L, R]$ into $N$ bins. Then the $i$-th bin's center $\mathbf{c}_{i}$ is $L + \frac{(i-0.5)}{N} * (R-L)$. For each pixel, the model will predict $N$ Softmax scores $\{\mathbf{p}_{1}, \mathbf{p}_2,\ldots, \mathbf{p}_N\}$, referring to the probabilities over the $N$ bins. The final predicted depth value $\tilde{D}$ is calculated from the linear combination of Softmax scores and bin centers at this pixel, as:
\begin{equation}
\tilde{D}=\sum_{i=1}^N\mathbf{c}_i\mathbf{p}_i.
\end{equation}

\subsection{Overview}
\noindent
We propose a new Depth AutoRegressive (DAR) modeling approach to explore the potential of autoregressive models in dealing with the depth estimation task. We define the task as follows: Given an input RGB image $\mathcal{I} \in \mathbb{R}^{3\times H \times W}$, where $H$ and $W$ are the height and width of $\mathcal{I}$ respectively, predict the depth map $\tilde{\mathcal{D}}$ of $\mathcal{I}$. Our model predicts the depth maps of different scales $\{\mathcal{\tilde{D}}_{1}, \mathcal{\tilde{D}}_{2}, \ldots,\mathcal{\tilde{D}}_{K}\}$ progressively in an autoregressive process. That is, each depth map at step $k$ is conditioned by the previous predictions, as: 
\vspace{-1pt}
{\small 
\begin{equation}
p(\mathcal{\tilde{D}}_1,\mathcal{\tilde{D}}_2,\ldots,\mathcal{\tilde{D}}_K)=\!\prod_{k=1}^Kp_{\theta}(\mathcal{\tilde{D}}_k\mid \mathcal{\tilde{D}}_1,\mathcal{\tilde{D}}_2,\ldots,\mathcal{\tilde{D}}_{k-1}).
\end{equation}}
\vskip -0.5 em \noindent
Our proposed autoregressive model DAR involves optimizing $p_{\theta}(\mathcal{\tilde{D}}_k\mid \mathcal{\tilde{D}}_1,\mathcal{\tilde{D}}_2,\ldots,\mathcal{\tilde{D}}_{k-1})$ over a dataset, and finally predicts depth map $\hat{\mathcal{D}} = \mathcal{\tilde{D}}_{K}$. Fig.~\ref{fig2} shows an overview of our proposed DAR. DAR involves two autoregressive objectives with ordinal properties: resolution and granularity. The former, resolution autoregressive objective, aims to predict the depth map from low resolution to high resolution; the latter,  granularity autoregressive objective, aims to predict the depth map from coarse granularity to fine granularity. Specifically, DAR consists of four parts:
\vskip -1 em
\begin{itemize}
    \item \textbf{Image Encoder}: We apply an Image Encoder to extract RGB image features, which extracts image features into imaging tokens with latent representations.

    \item \textbf{DAR Transformer}: Our DAR Transformer can progressively predict different resolution token maps conditioned on extracted RGB imaging tokens via the patch-wise causal mask. 

    \item \textbf{Multiway Tree Bins (MTBin)}: We develop a Multiway Tree Bins strategy to transform the depth range of each pixel into different granularity (number) bins. 

    \item \textbf{Bins Injection}: Bins Injection utilizes the bin candidates' information to guide the refinement of the latent features of the depth token maps.
\end{itemize}

\begin{figure}[t]
\includegraphics[width=0.467\textwidth]{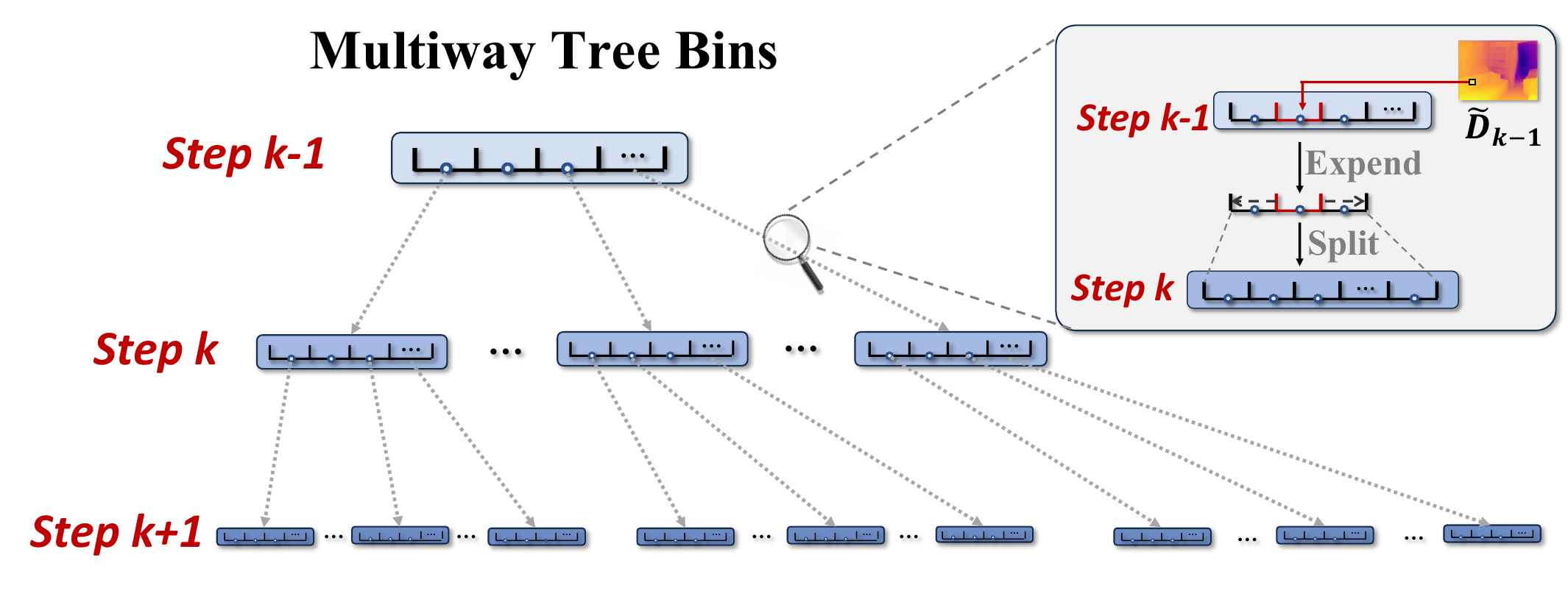}
\caption{A schematic diagram of the multiway tree bins strategy.}
\label{fig3}
\vskip -1 em
\end{figure}

\subsection{Resolution Autoregressive Objective}
This section elaborates on our DAR Transformer with the patch-wise causal attention mask for the resolution autoregressive objective.

\noindent
{\bf DAR Transformer.} Our DAR Transformer follows the vanilla architecture in~\cite{vaswani2017attention}, composed of Multi-headed Self-Attention (MSA), Layer Normalization (LN), and Multi-headed Cross-Attention (MCA) layers with residual connections, aiming to predict the logits sequence of different scale. Unlike text-to-image models that use class labels as the condition, depth estimation is mainly based on input RGB image features. Thus, we take the image features $X$ as the condition to control depth estimation. At each step $k$, we first upsample the token map $r_{k-1}$ of the previous step to the next resolution as the input token map $y_{in}^{k}$. The DAR Transformer takes $y_{in}^{k}$ as input query and sends it to the MSA and MCA layers to finally produce the logits $y_{out}^{k}$, where MSA takes the previous token maps $y_{in}^{1:k}$ to compute keys and values via the patch-wise causal attention mask, and MCA takes image features $X$ for attention calculation. We formulate the process at time step $k$ as:
\noindent
\begin{equation}
\centering
\begin{aligned}
y_{k}& = \text{Upsamping}(r_{k-1}), \\
y_{hidden}^{k}&=\text{LN}(\text{MSA}(y_{k}W_{Q};y_{1:k}W_{K};y_{1:k}W_{V}; \texttt{Mask})), \\
y_{out}^{k} &= \text{LN}(\text{MCA}(y_{hidden}^{k};X)), \\
\end{aligned}
\end{equation}
where $W_{Q}$, $W_{K}$, and $W_{V}$ are weight matrices for computing queries, keys, and values, and \texttt{Mask} denotes the patch-wise causal attention mask tailored for the next-resolution autoregressive paradigm (to be discussed below). The logits $y_{out}^k$ are then used to generate a latent token map $r_{k}$ of step $k$ guided by the Bins Injection module (to be discussed below). During this process, DAR can integrate previous knowledge, global image features, and bin candidate information to accommodate the detailed feature requirements for generating a higher-resolution depth map.

\noindent
{\bf Patch-wise Causal Attention Mask.} Note that to achieve ``next-resolution'' prediction, we adopt a patch-wise causal attention mask that treats the entire token map as a merged patch-wise token, as shown in Fig.~\ref{fig4}. This novel mask can ensure that each token of $y_{k}$ can interact only with its prefix tokens belonging to $y_{\leq k}$ and other tokens within $y_{k}$.

\subsection{Granularity Autoregressive Objective}
To achieve the autoregressive objective on granularity, we introduce two core modules: Multiway Tree Bins (MTBin) 
and Bins Injection.

\begin{table}[t]
\centering
\caption{Model sizes and architecture configurations of DAR. }
\scalebox{0.9}{
\begin{tabular}{lcccc}
\hline
Model     & Model Size & Layers & Hidden Size & Heads \\ \midrule
DAR-Small & 440M       & 5      & 1024        & 4     \\
DAR-Base  & 1B         & 7      & 1536        & 8     \\
DAR-Large & 2B       & 13     & 2560        & 12    \\ \bottomrule
\end{tabular}%
}
\label{par-tab}
\vskip -1 em
\end{table}

\noindent
{\bf Multiway Tree Bins (MTBin) Strategy.} This strategy aims to use the previous depth map's prediction to generate smaller bins for the next step's finer-grained depth predictions. Instead of using a fixed number of bins with equal-size intervals, MTBin recursively searches for more high-quality depth by progressively reducing the search range, as illustrated in Fig.~\ref{fig3}. Assuming that 
at step $k-1$, the depth range is divided into $N$ bins in the uniform space, as:
\vspace{-2 pt}
\begin{equation}
[\mathbf{b}_{k-1}^{1}, \mathbf{b}_{k-1}^{2},\ldots,\mathbf{b}_{k-1}^{N+1}], 
\end{equation}
\vskip -0.5 em \noindent
where $\mathbf{b}_{k-1}^{i}$ represents the left boundary of the $i$-th bin in step $k-1$, except $\mathbf{b}_{k-1}^{N+1}$ represents the right boundary of the $N$-th bin. Suppose the predicted depth $\tilde{D}_{k-1}(\mathbf{x})$ for a pixel $\mathbf{x}$ lies in the $t$-th target bin $[\mathbf{b}_{k-1}^{t}, \mathbf{b}_{k-1}^{t+1}]$, that is:
\vspace{-2 pt}
\begin{equation}
    \mathbf{b}_{k-1}^{t} \leq \tilde{D}_{k-1}(\mathbf{x}) \leq \mathbf{b}_{k-1}^{t+1}.
\end{equation}
\vskip -0.5 em \noindent
Then MTBin will recursively divide this bin into more fine-grained sub-bins, and update the depth range. However, the ground truth may fall outside of the target bin, due to depth prediction error. Thus, to maintain the model's error tolerance, the new depth range first  will be expanded to adjacent bins $[\mathbf{b}^{t-1}_{k-1}, \mathbf{b}^{t+2}_{k-1}]$, which is then split into sub-bins in the uniform space. This process can be formulated as:
\vspace{-1 pt}
\begin{align}
       L  &=  \frac{\mathbf{b}_{k-1}^{\min\{t+2, N+1\}} - \mathbf{b}_{k-1}^{\max\{t-1, 1\}}}{N}, \\
    \mathbf{b}_{k}^{i} &= \mathbf{b}_{k-1}^{t-1} + (i-1)\cdot L, \ i = 1, 2,\ldots, N, 
\end{align}
\vskip -0.5 em \noindent
where $\min\{t+2, N+1\}$ and $\max\{t-1, 1\}$ are for avoiding out-of-right/left boundaries. Since this splitting process looks like a multiway tree, we call this concept \textit{Multiway Tree Bins}. Each pixel's decision process is unique, ranging from coarse to fine granularity progressively. These sub-bins will serve as new bin candidates to further guide the modeling of depth features via \textit{Bins Injection} (discussed below) and perform the linear combination with the Softmax values of the token maps predicted by the model to obtain finer-grained depth maps. Specifically, we can take the bin centers as the depth candidates at step $k$, formulated as:
\vspace{-1 pt}
\begin{equation}
    \mathbf{c}_{k}^{i} = \frac{\mathbf{b}_{k}^{i}+\mathbf{b}_{k}^{i+1}}{2}, i = 1, 2,\ldots, N,
\end{equation}
where $\mathbf{c}_{\mathbf{i}}$ represents the $i$-th bin center (depth candidate). When we obtain the per-pixel Softmax values of $r_{k}$, i.e., the probabilistic distribution $\mathbf{p}_{k}$ associated with depth candidates, we compute the final depth via a linear combination:
\vspace{-1 pt}
\begin{equation}
    \tilde{D}_{k}(\mathbf{x}) = \sum_{i=1}^{N}\mathbf{c}_{k}^{i}\cdot\mathbf{p}_{k}^{i}\left(\mathbf{x}\right),
\end{equation}
\vskip -0.5 em \noindent
where $\tilde{D}_{k}(\mathbf{x})$ and $\mathbf{p}_{k}^{i}(\mathbf{x})$ represent the predicted depth and $i$-th depth candidate probability of pixel $\mathbf{x}$ at autoregressive step $k$, respectively. Notably, at first we initialize the full depth range $[d_{min},d_{max}]$ into $N$ bins in the uniform space:
\vspace{-1 pt}
\begin{equation}
    \mathbf{b}_{1}^{i} = d_{min} + (i - 1)\cdot L, \ i = 1, 2,\ldots, N,
\end{equation}
\vskip -0.5 em \noindent
where $\mathbf{b}_{1}^{i}$ represents the left boundary of the $i$-th bin at step 1, $N$ is 16 by default, [$d_{min}$, $d_{max}$] are [0.1, 10] and [0.1, 80] for \texttt{NYU Depth V2} and \texttt{KITTI}, respectively, and $L$ denotes the bin width that is equal to $\frac{d_{max}-d_{min}}{N}$.

\begin{table*}[t]
\centering
\caption{Results on the indoor \texttt{NYU Depth v2} dataset. \dag \ Data-driven methods which use a pretrained encoder on a large amount of data. }
\scalebox{0.83}{
\begin{tabular}{@{\quad}l@{\quad}@{\quad}l@{\quad}c@{\quad}@{\quad}ccccccc}
\toprule
     \multirow{2}{*}{Method} & \multirow{2}{*}{Encoder} & \multirow{2}{*}{Model Size} & 
    \multicolumn{4}{c}{Lower is better $\downarrow$} 
    &  \multicolumn{3}{c}{Higher is better $\uparrow$}
    \\
    \cmidrule(r){4-7} \cmidrule(r){8-10}  & & & Abs Rel & RMSE & Sq Rel & log$_{10}$ 
    & $\delta_{1}$ & $\delta_{2}$ & $\delta_{3}$
    \\ \midrule
    

Eigen et al.~\cite{eigen2014depth} &         ResNet-101  &  45M & 0.158              & 0.641     &    -    &  -       & 0.769   &  0.950  &  0.988  \\ 
DORN~\cite{fu2018deep} &           ResNet-101    & 45M &  0.115      &  0.509    &   -      &    0.051       &  0.828  &  0.965    &  0.992  \\ 
BTS~\cite{lee2019big}  & DenseNet-161  & 20M &  0.110       &  0.392    &   0.066   &     0.047        & 0.885    &  0.978   &  0.994  \\ 
AdaBins~\cite{bhat2021adabins} & E-B5+mini-ViT & 40M &   0.103   &    0.364     &    -    &   0.044      & 0.903    &   0.984 &  0.997   \\ 
P3Depth~\cite{patil2022p3depth} & ResNet-101                     &   45M           &   0.104      & 0.356      &    -     &  0.043        & 0.904   &  0.988  &  0.998    \\ 
DPT~\cite{ranftl2021vision} &    VIT-L   & 343M         &   0.110       &   0.357    &         &    0.045      &  0.904   & 0.988   &  0.998   \\ 
LocalBins~\cite{bhat2022localbins} &           E-B5 & 30M          &   0.099      &  0.357    &   -     &   0.042      & 0.907   &  0.987  & 0.998   \\ 
NeWCRFs~\cite{yuan2022neural} & Swin-Large  & 270M  &   0.095      &   0.334     &  0.045       &    0.041      &  0.922  &  0.992    &  0.998  \\ 
BinsFormer~\cite{li2024binsformer}                     &   Swin-Large     & 380M      &   0.094      &  0.330    &   -     &    0.040     &  0.925  & 0.991   & 0.997   \\ 
PixelFormer~\cite{agarwal2023attention}                  &   Swin-Large   & 365M         &   0.090      &  0.322    &  -      &   0.039      &  0.929  &  0.991  &  0.998  \\ 
VA-DepthNet~\cite{liu2023va}                       & Swin-Large   & 262M       &   0.086      &  0.304    &   0.043     &   0.039      &  0.929  &  0.991  &  0.998 \\ 
IEBins~\cite{shao2024iebins} &     Swin-Large  & 273M      &   0.087      & 0.314     &   0.040     &     0.038    &  0.936  & 0.992   & 0.998   \\ 
NDDepth~\cite{shao2023nddepth}                &    Swin-Large    &  393M     &   0.087      &  0.311    &   0.041     &  0.038       & 0.936   & 0.991   &  0.998  \\ 
DCDepth\cite{wang2024dcdepth}               &    Swin-Large    &  259M     &   0.085      &  0.304    &   0.039     &  0.037       & 0.940   & 0.992   &  0.998  \\ 
WorDepth~\cite{zeng2024wordepth}           &   Swin-Large  & 350M        &         0.088           &    0.317     &     -        &   0.038      &  0.932  & 0.992   & 0.998   \\
VPD ~\cite{zhao2023unleashing}               &   ViT-L   & 600M   &  0.069              &  0.254       &  0.030     &   0.027      &    0.964      &  0.995   &  0.999    \\ 
EcoDepth~\cite{patni2024ecodepth}               &   ViT-L & 954M     &  0.059              &  0.218       &  0.013     &   0.026      &    0.978      &  0.997   &  0.999    \\ \hline
ZoeDepth\dag ~\cite{bhat2023zoedepth}        & ViT-L  & 343M  &   0.077        &   0.282    &   -     &    0.033     &  0.951  &   0.994  &  0.999   \\       
Depth Anything\dag~\cite{yang2024depth}              & ViT-L & 343M   &      0.063       &  0.235   &    0.020    &  0.026         &  0.975  & 0.997   & 0.999   \\  \hline 

 DAR-Small             &       ViT-L    & 440M      & {\bf 0.059}     &  {\bf 0.217}    & {\bf    0.013}    &   {\bf  0.026}     &  {\bf 
 0.979}  & {\bf  0.997}  & {\bf  0.999}  \\ 
  DAR-Base             &       ViT-L    & 1.0B      & {\bf 0.058}     &  {\bf 0.214}    & {\bf    0.013}    &   {\bf  0.026}     &  {\bf 
 0.980}  & {\bf  0.997}  & {\bf  0.999}  \\ 
 DAR-Large                     &       ViT-L    & 2.0B     & {\bf 0.056}     &  {\bf 0.205}    & {\bf    0.011}    &   {\bf  0.024}     &  {\bf 
 0.982}  & {\bf  0.998}  & {\bf  1.000}  \\ \hline
\end{tabular}%
}
\label{tab1}
\end{table*}

\begin{figure*}[t]
\centering
\includegraphics[width=0.82\textwidth]{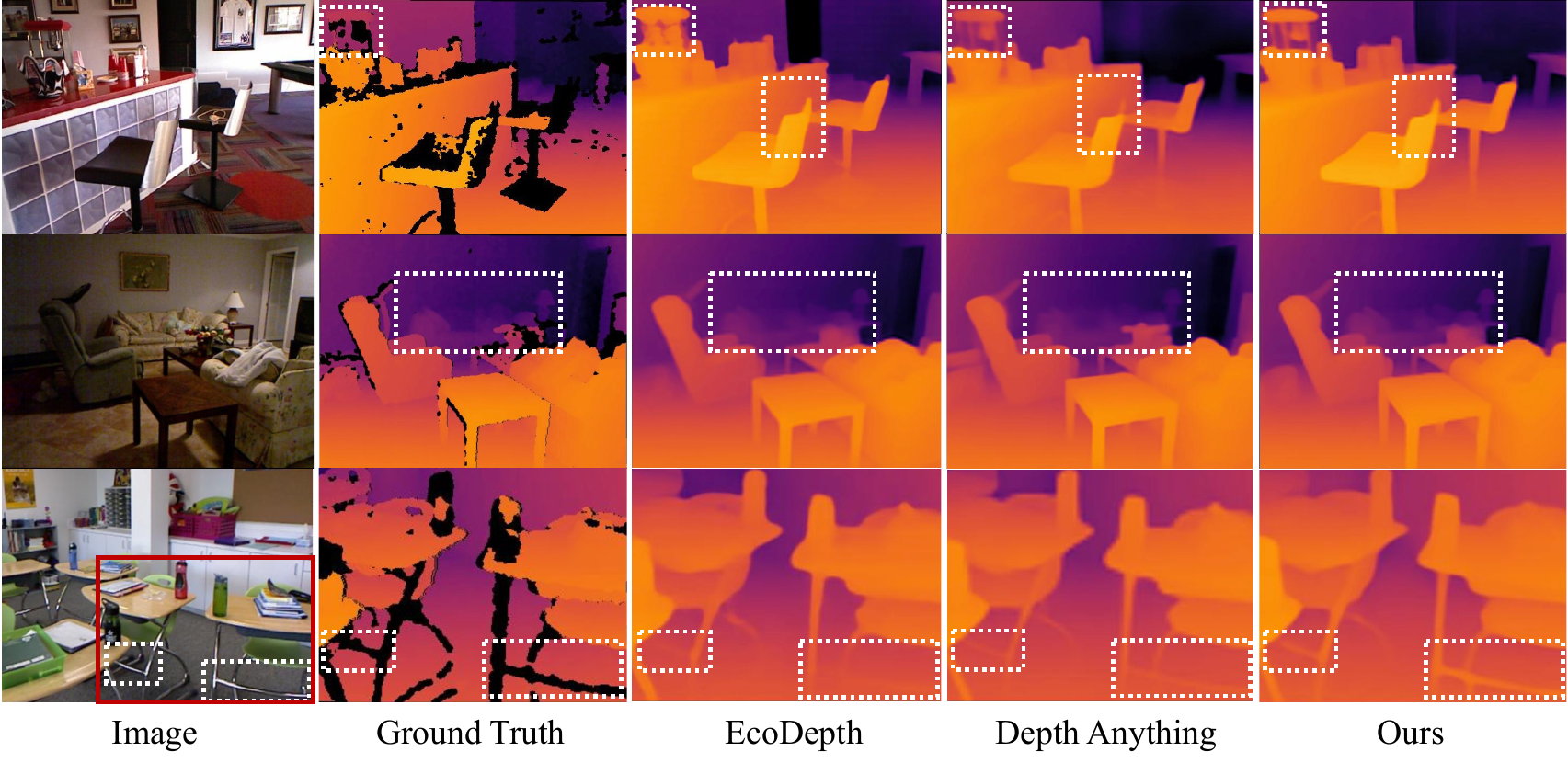}
\caption{Qualitative results on the \texttt{NYU Depth V2} dataset. A brighter color denotes a closer distance.}
\label{figq1}
\vskip -1 em
\end{figure*}

\begin{table*}[h]
\centering
\caption{Results on the outdoor \texttt{KITTI} dataset. \dag \ Data-driven methods which use a pretrained backbone on a large amount of data.}
\scalebox{0.83}{
\begin{tabular}{@{\quad}l@{\quad}@{\quad}l@{\quad}c@{\quad}@{\quad}ccccccc}
\toprule
     \multirow{2}{*}{Method} & \multirow{2}{*}{Encoder} & \multirow{2}{*}{Model Size} & 
    \multicolumn{4}{c}{Lower is better $\downarrow$} 
    &  \multicolumn{3}{c}{Higher is better $\uparrow$}
    \\
    \cmidrule(r){4-7} \cmidrule(r){8-10}  & & & Abs Rel & RMSE & Sq Rel & RMSE$_{log}$
    & $\delta_{1}$ & $\delta_{2}$ & $\delta_{3}$
    \\ \midrule
    

Eigen et al.~\cite{eigen2014depth} &  ResNet-101  &  45M    & 0.203              & 6.307     &    1.517    &  0.282       & 0.702   &  0.898  &  0.967  \\ 
DORN~\cite{fu2018deep} & ResNet-101    & 45M     &   0.072           &  2.727    &   0.307      &    0.120       &  0.932  &   0.984    &  0.994  \\ 
BTS~\cite{lee2019big} & DenseNet-161  & 20M &   0.059           &  2.756   &   0.241  &     0.096       & 0.956    &  0.993   &   0.998 \\ 
AdaBins~\cite{bhat2021adabins} & E-B5+mini-ViT & 40M &    0.067       &    2.960     &    0.190    &   0.088      & 0.949     &   0.992  &  0.998   \\ 
DPT~\cite{ranftl2021vision} &   VIT-L   & 343M              &   0.060             &   2.573    &    -    &   0.092      &  0.959   & 0.995   &  0.996   \\ 
P3Depth~\cite{patil2022p3depth} & ResNet-101    & 45M &         0.071       & 2.842       &   0.270     &  0.103          & 0.953    &  0.993  &  0.998    \\ 
NeWCRFs~\cite{yuan2022neural} & Swin-Large  & 270M  &   0.052          &   2.129    &  0.155       &    0.079    &  0.974  &  0.997   &  0.999   \\ 
BinsFormer~\cite{li2024binsformer}                    &   Swin-Large    & 380M        &   0.052         &  2.098   &   0.151  &  0.079        &  0.974  & 0.997    & 0.999   \\ 
PixelFormer~\cite{agarwal2023attention}      &   Swin-Large    & 365M        &   0.051         &  2.081   &   0.149  &  0.077        &  0.976  & 0.997    & 0.999   \\ 
VA-DepthNet~\cite{liu2023va} & Swin-Large   & 262M          &   0.050      &  2.093    &   0.148     &   0.076      &  0.977  &  0.997  &  0.999 \\ 
IEBins~\cite{shao2024iebins} &     Swin-Large  & 273M        &   0.050      & 2.011     &   0.142     &   -      &  0.978  & 0.998   & 0.999   \\ 
iDisc~\cite{piccinelli2023idisc}  &  Swin-Large     & 263M      &   0.050      &  2.067    &   0.145     & 0.077       & 0.977  &  0.997  &  0.999  \\ 
DCDepth~\cite{wang2024dcdepth} & Swin-Large    &  259M            &   0.051      &  2.044    &   0.145     & 0.076       & 0.977  &  0.997  &  0.999 \\  
WorDepth~\cite{zeng2024wordepth}             &   Swin-Large  &  350M     &         0.049           &    2.039     &     -        &   0.074      &  0.979      &  0.998   &  0.999   \\
EcoDepth~\cite{patni2024ecodepth} & ViT-L & 954M      &  0.048              &  2.039       &  0.139     &   0.074      &    0.979      &  0.998   &  1.000    \\ \hline
ZoeDepth\dag ~\cite{bhat2023zoedepth}    & ViT-L  & 343M    &   0.054            &   2.440    &   0.189    &    0.083     &  0.977   &   0.996  &  0.999   \\       
Depth Anything\dag~\cite{yang2024depth}  & ViT-L & 343M     &      0.046       &  1.896  &    -    &  0.069        &  0.982   & 0.998   & 1.000   \\  \hline 
DAR-small & ViT-L    & 440M          & {\bf 0.046}     &  {\bf 1.839}    & {\bf    0.115}    &   {\bf  0.069}     &  {\bf 
 0.984}  & {\bf  0.999}  & {\bf  1.000}  \\ 
DAR-Base &  ViT-L    & 1.0B          & {\bf 0.046}     &  {\bf 1.823}    & {\bf    0.114}    &   {\bf  0.069}     &  {\bf 
 0.985}  & {\bf  0.999}  & {\bf  1.000}  \\ 
DAR-Large & ViT-L    & 2.0B         & {\bf 0.044}     &  {\bf 1.799}    & {\bf    0.110}    &   {\bf   0.067}     &  {\bf 
 0.986}  & {\bf  0.999}  & {\bf  1.000}  \\ \hline
\end{tabular}%
}
\label{tab2}
\end{table*}

\begin{figure*}[t]
\centering
\includegraphics[width=0.82\textwidth]{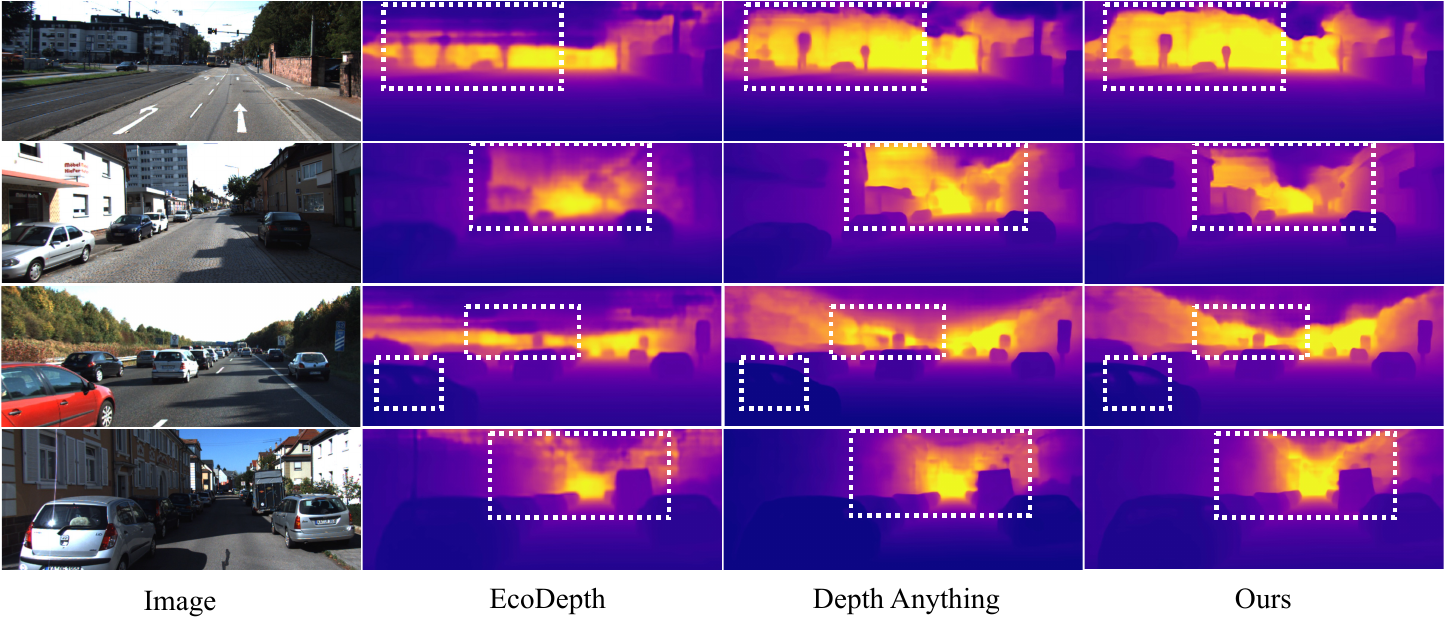}
\caption{Qualitative results on the \texttt{KITTI} dataset. A darker color denotes a closer distance.}
\label{figq2}
\end{figure*}

\noindent 
{\bf Bins Injection.} Bins Injection module aims to utilize the new valid depth range and bin candidates to guide the modeling of depth features.  First, we project the depth candidates $\mathbf{c}^{k}$ into the feature space via $3 \times 3$ convolutional layers. Then, the obtained bin features $\mathbf{f}_{bin}^{k}$ as the context, are used to further guide the output features of the DAR Transformer via a ConvGRU~\cite{teed2020raft} module. We formulate this process as:
\vspace{-1 pt}
\begin{align}
    \mathbf{f}_{bin}^{k} &= \text{Conv}_{3 \times 3} (\mathbf{c}^{k}), \\ 
    r_{k} &= \text{ConvGRU}(\mathbf{y}_{out}^{k};\mathbf{f}_{bin}^{k}).
\end{align}
\vskip -1 em

\subsection{Other Details}

\noindent
{\bf Image Encoder.} To ensure a fair comparison with existing methods, we choose ViT as the Image Encoder (which is the same as Depth Anything). The conditioned image feature tokens are obtained by aggregating the feature maps from different layers of the Image Encoder, by bringing them all to 1/8 resolution of the input image, resulting in a token map of size $1536 \times H/8 \times W/8$, where $H$ and $W$ are the height and width of the input RGB image.

\noindent
{\bf Loss Function.} Since the ground-truth depth map has missing values, we cannot resize the ground-truth to different resolutions. Thus, we upsample all the predicted depth maps to the same size of the ground-truth, and calculate and leverage the scaled Scale-Invariant Loss:
{\small 
\begin{equation}
\mathcal{L} =\sum_{k=1}^K\alpha\sqrt{\frac1{|\mathbf{T}|}\sum_{\mathbf{x \in T}}\left(\mathbf{g}_{k}\left(\mathbf{x}\right)\right)^2-\frac\beta{|\mathbf{T}|^2}\left(\sum_{\mathbf{x \in T}}\mathbf{g}_{k}\left(\mathbf{x}\right)\right)^2}\!, 
\end{equation}
}
where $\mathbf{g}_{k}(\mathbf{x}) = \text{log} \tilde{\mathcal{D}}_{k}(\mathbf{x}) - \text{log}\mathcal{D}_{gt}(\mathbf{x})$, $K$ is the maximum number of steps and is set to 5, $T$ represents the set of pixels having valid ground-truth values, and $\alpha$ and $\beta$ are set to 10 and 0.85 based on~\cite{lee2019big}.

\section{Experiments}

\subsection{Datasets and Evaluation Metrics}

\noindent
{\bf Scaling Up Setting.}
Following~\cite{kaplan2020scaling,henighan2020scaling,hoffmann2022training,achiam2023gpt}, we scale the model up with different sizes of the DAR Transformer. The configurations are shown in Table~\ref{par-tab}. 

\noindent
{\bf Implementation Details.} 
Our model is implemented on the PyTorch platform. For optimization, the AdamW optimizer~\cite{kingma2014adam} is used with an initial learning rate of $3 \times 10^{-5}$. We first linearly increase the learning rate to $5 \times 10^{-4}$, and then linearly decrease it across the training iterations. The mini-batch size is 16. We train our model for 25 epochs for both the \texttt{KITTI} and \texttt{NYU Depth v2} datasets. In each step, the number of new bins is $N=16$. For the DAR-Base model, each epoch takes $\sim$30 minutes to train using 8 NVIDIA A100 GPUs.

\noindent
{\bf Datasets.} We conduct experiments on three benchmark datasets including \texttt{NYU Depth V2}~\cite{silberman2012indoor}, \texttt{KITTI}~\cite{geiger2012we}, and \texttt{SUN RGB-D}~\cite{song2015sun}. {\bf (a) \texttt{NYU Depth V2}} is a widely-used benchmark dataset that covers indoor scenes with depth values ranging from 0 to 10 meters. We follow the train-test split in~\cite{lee2019monocular}, which uses 24,231 images for training and 654 images for testing. The ground truth depth maps were obtained using a structured light sensor with a resolution of $640 \times 480$. {\bf (b) \texttt{KITTI}} is a widely-used outdoor benchmark dataset, containing images with depth values ranging from 0 to 80 meters.  The official split provides 42,949 images for training, 1,000 for validation, and 500 for testing, with resolution $352 \times 1216$. {\bf (c) \texttt{SUN RGB-D}}: We preprocess its images to 480 × 640 resolution for consistency. The depth values range from 0 to 10 meters. We use only the official test set (5050 images) for zero-shot evaluation.

\begin{table}[t]
\centering
\caption{Zero-shot generalization to the SUN RGB-D dataset~\cite{song2015sun}. The models are trained on \texttt{NYU Depth V2} and tested on SUN RGB-D without fine-tuning.}
\scalebox{0.74}{
\begin{tabular}{lcccccc}
\toprule
     \multirow{2}{*}{Method}  & 
    \multicolumn{3}{c}{Lower is better $\downarrow$} 
    &  \multicolumn{3}{c}{Higher is better $\uparrow$}
    \\
    \cmidrule(r){2-4} \cmidrule(r){5-7}  & Abs Rel & RMSE  & log$_{10}$ 
    & $\delta_{1}$ & $\delta_{2}$ & $\delta_{3}$
    \\ \midrule
BinsFormer~\cite{li2024binsformer} &   0.143           &   0.421         &    0.061    &  0.805   &  0.963    &  0.990  \\
IEBins~\cite{shao2024iebins} &   0.135          &   0.405         &    0.059     &  0.822   &  0.971   &  0.993    \\
VPD~\cite{zhao2023unleashing}  &   0.121          &   0.355        &    0.045    &  0.861  &  0.980   &  0.995   \\ 
Depth Anything~\cite{yang2024depth}  &   0.119          &   0.346       &    0.043    &  0.864  &  0.981   &  0.995 \\ \hline
DAR  &   {\bf 0.112}         &  {\bf 0.319}       &  {\bf  0.040}  & {\bf 0.885} & {\bf 0.985}  & {\bf 0.996}  \\
\bottomrule
\end{tabular}%
}
\label{tab3}
\vskip -1 em
\end{table}

\subsection{Comparisons with Previous Methods}

\noindent
{\bf Quantitative Results on \texttt{NYU Depth V2}.} We report the results in Table~\ref{tab1}. As one can see, DAR with the same pretrained backbone and similar model size performs better than and above all the known methods. By scaling up, DAR-Large establishes a new SOTA performance in all the metrics, attaining 0.205 RMSE and 0.982 $\delta_{1}$ on \texttt{NYU Depth v2}, largely outperforming existing methods.

\noindent {\bf Qualitative Results on \texttt{NYU Depth V2}.} In Fig.~\ref{figq1}, we present some qualitative comparison on the dataset. First, one can observe that our model performs better in depth estimation at the boundaries of the objects, making it more coherent and smooth (e.g., the back of the chair and the long-distant objects). This is helped by our autoregressive progressive paradigm, which maintains a coherent and smooth depth estimation when using previous predictions for the next-step, more refined prediction. Second, our DAR is much more accurate when estimating the depths of small and thin objects or long-distant visually relatively small objects, like the poles under the chair. These observations further demonstrate the superiority of our DAR.

\noindent
{\bf Quantitative Results on \texttt{KITTI}.} 
To further demonstrate the superiority of our proposed DAR in outdoor scenarios, we report results on the \texttt{KITTI} dataset in Table~\ref{tab2}. Compared to the SOTA self-supervised model Depth Anything, our DAR-small with a similar model size achieves better performance, suggesting the superiority and potential of our proposed DAR. Compared to the SOTA supervised model Ecodepth, our DAR-small has smaller model sizes and performs better. Notably, the scaled-up version DAR-Large establishes a new SOTA performance, which outperforms the current SOTA (Depth Anything) in all the metrics (in particular, by 5\% in RMSE). 

\noindent {\bf Qualitative Results on \texttt{KITTI}.} In Fig.~\ref{figq2}, we present some qualitative comparison on the \texttt{KITTI} dataset. One can observe that DAR preserves fine-grained boundary details and generates more continuous depth values, further demonstrating the effectiveness of our new AR-based framework.

\subsection{Zero-shot Generalization}
Unlike the SOTA zero-shot transfer method (Depth Anything~\cite{yang2024depth}), which requires pretraining on a large amount of data (61M) for effective performance, we show that our model generalizes well when trained only on a single \texttt{NYU Depth v2} dataset. Table~\ref{tab3} shows the quantitative results. As one can see, DAR achieves decent results on unseen datasets, substantiating the generalization ability of DAR.

\begin{table}[t]
\centering
\caption{Ablation study of DAR. ``BI'': The Bins Injection module.}
\scalebox{0.79}{
\begin{tabular}{lcccc}
\toprule
Method                             & Param. & Abs Rel $\downarrow$ & RMSE $\downarrow$ & d1 $\uparrow$ \\ \midrule
Baseline + Transformer             &  420M &  0.063      & 0.229     & 0.976   \\ 
Baseline + MTBins + BI             &  363M       &  0.061      &   0.220   &  0.978   \\
Baseline + DAR                     &  440M  &    0.059    &  0.217    & 0.979   \\ 
Baseline + DAR + Scale Up          &  2.0B   &  0.056      &  0.205    & 0.982   \\ \bottomrule
\end{tabular}
}
\label{taba}
\end{table}

\subsection{Ablation Study}
We conduct a comprehensive ablation study to examine the effectiveness of the two sub-AR objectives with their respective components and scale up. All the experiments presented in this section are conducted on \texttt{NYU-Depth-V2}. We choose the backbone of Depth Anything as the baseline model. The comparison results are reported in Table~\ref{taba}. Observe that each sub-AR objective and its components improve the baseline performance, supporting our hypothesis that the AR model is an effective monocular depth estimator. Furthermore, by scaling up the model size to 2.0B, DAR achieves the best performance in all the metrics, demonstrating the strong scalability of DAR. We also conduct ablation study on the bin number $N$, and show qualitative results of each step of DAR in the Supplemental Materials.

\subsection{Limitations}

We need to acknowledge the following limitations. First, we apply a multi-step progressive paradigm to predict depths, which is smoother and more continuous, but consequently, it may blur the boundaries and reduce sharpness. Second, since we use the autoregressive Transformer, the model parameter number of DAR is relatively high, especially when scaling the model size up. But, we believe that further improvements in the complexity-accuracy trade-off of DAR can be achieved through large model distillation or lightweight AR foundation model design techniques.

\section{Conclusions}
In this paper, we presented a novel depth autoregressive (DAR) modeling approach to introduce scalability and generalization to an AR-based framework for monocular depth estimation (MDE). Our key idea is to transform the MDE task into two parallel autoregressive objectives, resolution and granularity, inspired by the insight of ordered properties of MDE in these two aspects. Further, we proposed the DAR Transformer and Multiway Tree Bins strategy to achieve these two autoregressive objectives, and connect them by the proposed Bin Injection module. We demonstrated the effectiveness of our approach on several benchmark datasets, and showed that it outperforms SOTA methods by a significant margin. Our proposed DAR also offers a promising way to integrate autoregressive depth estimation into existing autoregressive-based foundation models.





{\small
\bibliographystyle{ieee_fullname}
\bibliography{egbib}

\begin{thebibliography}{10}\itemsep=-1pt

\bibitem{achiam2023gpt}
Josh Achiam, Steven Adler, Sandhini Agarwal, Lama Ahmad, Ilge Akkaya, Florencia~Leoni Aleman, Diogo Almeida, Janko Altenschmidt, Sam Altman, Shyamal Anadkat, et~al.
\newblock {GPT-4} technical report.
\newblock {\em arXiv preprint arXiv:2303.08774}, 2023.

\bibitem{agarwal2023attention}
Ashutosh Agarwal and Chetan Arora.
\newblock Attention attention everywhere: Monocular depth prediction with skip attention.
\newblock In {\em Proceedings of the IEEE/CVF Winter Conference on Applications of Computer Vision}, pages 5861--5870, 2023.

\bibitem{aich2021bidirectional}
Shubhra Aich, Jean Marie~Uwabeza Vianney, Md~Amirul Islam, and Mannat Kaur~Bingbing Liu.
\newblock Bidirectional attention network for monocular depth estimation.
\newblock In {\em 2021 IEEE International Conference on Robotics and Automation (ICRA)}, pages 11746--11752. IEEE, 2021.

\bibitem{bhat2021adabins}
Shariq~Farooq Bhat, Ibraheem Alhashim, and Peter Wonka.
\newblock {AdaBins}: Depth estimation using adaptive bins.
\newblock In {\em Proceedings of the IEEE/CVF Conference on Computer Vision and Pattern Recognition}, pages 4009--4018, 2021.

\bibitem{bhat2022localbins}
Shariq~Farooq Bhat, Ibraheem Alhashim, and Peter Wonka.
\newblock {LocalBins}: Improving depth estimation by learning local distributions.
\newblock In {\em European Conference on Computer Vision}, pages 480--496. Springer, 2022.

\bibitem{bhat2023zoedepth}
Shariq~Farooq Bhat, Reiner Birkl, Diana Wofk, Peter Wonka, and Matthias M{\"u}ller.
\newblock {ZoeDepth}: Zero-shot transfer by combining relative and metric depth.
\newblock {\em arXiv preprint arXiv:2302.12288}, 2023.

\bibitem{brown2020language}
Tom~B Brown.
\newblock Language models are few-shot learners.
\newblock {\em arXiv preprint arXiv:2005.14165}, 2020.

\bibitem{chatterjee2024robustness}
Agneet Chatterjee, Tejas Gokhale, Chitta Baral, and Yezhou Yang.
\newblock On the robustness of language guidance for low-level vision tasks: Findings from depth estimation.
\newblock In {\em Proceedings of the IEEE/CVF Conference on Computer Vision and Pattern Recognition}, pages 2794--2803, 2024.

\bibitem{chen2019towards}
Po-Yi Chen, Alexander~H Liu, Yen-Cheng Liu, and Yu-Chiang~Frank Wang.
\newblock Towards scene understanding: Unsupervised monocular depth estimation with semantic-aware representation.
\newblock In {\em Proceedings of the IEEE/CVF Conference on Computer Vision and Pattern Recognition}, pages 2624--2632, 2019.

\bibitem{chen2021pix2seq}
Ting Chen, Saurabh Saxena, Lala Li, David~J Fleet, and Geoffrey Hinton.
\newblock {Pix2Seq}: A language modeling framework for object detection.
\newblock {\em arXiv preprint arXiv:2109.10852}, 2021.

\bibitem{duan2023diffusiondepth}
Yiqun Duan, Xianda Guo, and Zheng Zhu.
\newblock {DiffusionDepth}: Diffusion denoising approach for monocular depth estimation.
\newblock {\em arXiv preprint arXiv:2303.05021}, 2023.

\bibitem{eigen2014depth}
David Eigen, Christian Puhrsch, and Rob Fergus.
\newblock Depth map prediction from a single image using a multi-scale deep network.
\newblock {\em NIPS}, 27, 2014.

\bibitem{esser2021taming}
Patrick Esser, Robin Rombach, and Bjorn Ommer.
\newblock Taming {Transformers} for high-resolution image synthesis.
\newblock In {\em Proceedings of the IEEE/CVF Conference on Computer Vision and Pattern Recognition}, pages 12873--12883, 2021.

\bibitem{fu2018deep}
Huan Fu, Mingming Gong, Chaohui Wang, Kayhan Batmanghelich, and Dacheng Tao.
\newblock Deep ordinal regression network for monocular depth estimation.
\newblock In {\em Proceedings of the IEEE Conference on Computer Vision and Pattern Recognition}, pages 2002--2011, 2018.

\bibitem{geiger2012we}
Andreas Geiger, Philip Lenz, and Raquel Urtasun.
\newblock Are we ready for autonomous driving? the {KITTI} vision benchmark suite.
\newblock In {\em 2012 IEEE Conference on Computer Vision and Pattern Recognition}, pages 3354--3361. IEEE, 2012.

\bibitem{harrell2012regression}
Frank~E Harrell.
\newblock Regression modeling strategies.
\newblock {\em R Package Version}, pages 6--2, 2012.

\bibitem{henighan2020scaling}
Tom Henighan, Jared Kaplan, Mor Katz, Mark Chen, Christopher Hesse, Jacob Jackson, Heewoo Jun, Tom~B Brown, Prafulla Dhariwal, Scott Gray, et~al.
\newblock Scaling laws for autoregressive generative modeling.
\newblock {\em arXiv preprint arXiv:2010.14701}, 2020.

\bibitem{herbrich1999support}
Ralf Herbrich, Thore Graepel, and Klaus Obermayer.
\newblock Support vector learning for ordinal regression.
\newblock 1999.

\bibitem{hoffmann2022training}
Jordan Hoffmann, Sebastian Borgeaud, Arthur Mensch, Elena Buchatskaya, Trevor Cai, Eliza Rutherford, Diego de~Las Casas, Lisa~Anne Hendricks, Johannes Welbl, Aidan Clark, et~al.
\newblock Training compute-optimal large language models.
\newblock {\em arXiv preprint arXiv:2203.15556}, 2022.

\bibitem{hoiem2007recovering}
Derek Hoiem, Alexei~A Efros, and Martial Hebert.
\newblock Recovering surface layout from an image.
\newblock {\em International Journal of Computer Vision}, 75:151--172, 2007.

\bibitem{izadi2011kinectfusion}
Shahram Izadi, David Kim, Otmar Hilliges, David Molyneaux, Richard Newcombe, Pushmeet Kohli, Jamie Shotton, Steve Hodges, Dustin Freeman, Andrew Davison, et~al.
\newblock {KinectFusion}: Real-time {3D} reconstruction and interaction using a moving depth camera.
\newblock In {\em Proceedings of the 24th annual ACM symposium on User Interface Software and Technology}, pages 559--568, 2011.

\bibitem{ji2023ddp}
Yuanfeng Ji, Zhe Chen, Enze Xie, Lanqing Hong, Xihui Liu, Zhaoqiang Liu, Tong Lu, Zhenguo Li, and Ping Luo.
\newblock {DDP}: Diffusion model for dense visual prediction.
\newblock In {\em Proceedings of the IEEE/CVF International Conference on Computer Vision}, pages 21741--21752, 2023.

\bibitem{jia2023object}
Weibin Jia, Wenjie Zhao, Zhihuan Song, and Zhengguo Li.
\newblock Object servoing of differential-drive service robots using switched control.
\newblock {\em Journal of Control and Decision}, 10(3):314--325, 2023.

\bibitem{kaplan2020scaling}
Jared Kaplan, Sam McCandlish, Tom Henighan, Tom~B Brown, Benjamin Chess, Rewon Child, Scott Gray, Alec Radford, Jeffrey Wu, and Dario Amodei.
\newblock Scaling laws for neural language models.
\newblock {\em arXiv preprint arXiv:2001.08361}, 2020.

\bibitem{ke2024repurposing}
Bingxin Ke, Anton Obukhov, Shengyu Huang, Nando Metzger, Rodrigo~Caye Daudt, and Konrad Schindler.
\newblock Repurposing diffusion-based image generators for monocular depth estimation.
\newblock In {\em Proceedings of the IEEE/CVF Conference on Computer Vision and Pattern Recognition}, pages 9492--9502, 2024.

\bibitem{kingma2014adam}
Diederik~P Kingma.
\newblock Adam: A method for stochastic optimization.
\newblock {\em arXiv preprint arXiv:1412.6980}, 2014.

\bibitem{laina2016deeper}
Iro Laina, Christian Rupprecht, Vasileios Belagiannis, Federico Tombari, and Nassir Navab.
\newblock Deeper depth prediction with fully convolutional residual networks.
\newblock In {\em 2016 Fourth international conference on 3D vision (3DV)}, pages 239--248. IEEE, 2016.

\bibitem{lee2022autoregressive}
Doyup Lee, Chiheon Kim, Saehoon Kim, Minsu Cho, and Wook-Shin Han.
\newblock Autoregressive image generation using residual quantization.
\newblock In {\em Proceedings of the IEEE/CVF Conference on Computer Vision and Pattern Recognition}, pages 11523--11532, 2022.

\bibitem{lee2019big}
Jin~Han Lee, Myung-Kyu Han, Dong~Wook Ko, and Il~Hong Suh.
\newblock From big to small: Multi-scale local planar guidance for monocular depth estimation.
\newblock {\em arXiv preprint arXiv:1907.10326}, 2019.

\bibitem{lee2019monocular}
Jae-Han Lee and Chang-Su Kim.
\newblock Monocular depth estimation using relative depth maps.
\newblock In {\em Proceedings of the IEEE/CVF Conference on Computer Vision and Pattern Recognition}, pages 9729--9738, 2019.

\bibitem{li2023depthformer}
Zhenyu Li, Zehui Chen, Xianming Liu, and Junjun Jiang.
\newblock {DepthFormer}: Exploiting long-range correlation and local information for accurate monocular depth estimation.
\newblock {\em Machine Intelligence Research}, 20(6):837--854, 2023.

\bibitem{li2024binsformer}
Zhenyu Li, Xuyang Wang, Xianming Liu, and Junjun Jiang.
\newblock {BinsFormer}: Revisiting adaptive bins for monocular depth estimation.
\newblock {\em IEEE Transactions on Image Processing}, 2024.

\bibitem{liu2023va}
Ce Liu, Suryansh Kumar, Shuhang Gu, Radu Timofte, and Luc Van~Gool.
\newblock {VA-DepthNet}: A variational approach to single image depth prediction.
\newblock {\em arXiv preprint arXiv:2302.06556}, 2023.

\bibitem{liu2008sift}
Ce Liu, Jenny Yuen, Antonio Torralba, Josef Sivic, and William~T Freeman.
\newblock {SIFT Flow}: Dense correspondence across different scenes.
\newblock In {\em Computer Vision--ECCV 2008: 10th European Conference on Computer Vision, Marseille, France, October 12-18, 2008, Proceedings, Part III 10}, pages 28--42. Springer, 2008.

\bibitem{liu2024visual}
Haotian Liu, Chunyuan Li, Qingyang Wu, and Yong~Jae Lee.
\newblock Visual instruction tuning.
\newblock {\em Advances in Neural Information Processing Systems}, 36, 2024.

\bibitem{liu2019dense}
Xingtong Liu, Ayushi Sinha, Masaru Ishii, Gregory~D Hager, Austin Reiter, Russell~H Taylor, and Mathias Unberath.
\newblock Dense depth estimation in monocular endoscopy with self-supervised learning methods.
\newblock {\em IEEE Transactions on Medical Imaging}, 39(5):1438--1447, 2019.

\bibitem{oney2020evaluation}
Seyda Oney, Nils Rodrigues, Michael Becher, Thomas Ertl, Guido Reina, Michael Sedlmair, and Daniel Weiskopf.
\newblock Evaluation of gaze depth estimation from eye tracking in augmented reality.
\newblock In {\em ACM Symposium on Eye Tracking Research and Applications}, pages 1--5, 2020.

\bibitem{patil2022p3depth}
Vaishakh Patil, Christos Sakaridis, Alexander Liniger, and Luc Van~Gool.
\newblock {P3Depth}: Monocular depth estimation with a piecewise planarity prior.
\newblock In {\em Proceedings of the IEEE/CVF Conference on Computer Vision and Pattern Recognition}, pages 1610--1621, 2022.

\bibitem{patni2024ecodepth}
Suraj Patni, Aradhye Agarwal, and Chetan Arora.
\newblock {ECoDepth}: Effective conditioning of diffusion models for monocular depth estimation.
\newblock In {\em Proceedings of the IEEE/CVF Conference on Computer Vision and Pattern Recognition}, pages 28285--28295, 2024.

\bibitem{piccinelli2023idisc}
Luigi Piccinelli, Christos Sakaridis, and Fisher Yu.
\newblock {iDisc}: Internal discretization for monocular depth estimation.
\newblock In {\em Proceedings of the IEEE/CVF Conference on Computer Vision and Pattern Recognition}, pages 21477--21487, 2023.

\bibitem{radford2018improving}
Alec Radford.
\newblock Improving language understanding by generative pre-training.
\newblock 2018.

\bibitem{radford2019language}
Alec Radford, Jeffrey Wu, Rewon Child, David Luan, Dario Amodei, Ilya Sutskever, et~al.
\newblock Language models are unsupervised multitask learners.
\newblock {\em OpenAI Blog}, 1(8):9, 2019.

\bibitem{ranftl2021vision}
Ren{\'e} Ranftl, Alexey Bochkovskiy, and Vladlen Koltun.
\newblock Vision {Transformers} for dense prediction.
\newblock In {\em Proceedings of the IEEE/CVF International Conference on Computer Vision}, pages 12179--12188, 2021.

\bibitem{ranftl2020towards}
Ren{\'e} Ranftl, Katrin Lasinger, David Hafner, Konrad Schindler, and Vladlen Koltun.
\newblock Towards robust monocular depth estimation: Mixing datasets for zero-shot cross-dataset transfer.
\newblock {\em IEEE Transactions on Pattern Analysis and Machine Intelligence}, 44(3):1623--1637, 2020.

\bibitem{razavi2019generating}
Ali Razavi, Aaron Van~den Oord, and Oriol Vinyals.
\newblock Generating diverse high-fidelity images with {VQ-VAE-2}.
\newblock {\em Advances in Neural Information Processing Systems}, 32, 2019.

\bibitem{saxena2007learning}
Ashutosh Saxena, Min Sun, and Andrew~Y Ng.
\newblock Learning 3-{D} scene structure from a single still image.
\newblock In {\em 2007 IEEE 11th International Conference on Computer Vision}, pages 1--8. IEEE, 2007.

\bibitem{saxena2023monocular}
Saurabh Saxena, Abhishek Kar, Mohammad Norouzi, and David~J Fleet.
\newblock Monocular depth estimation using diffusion models.
\newblock {\em arXiv preprint arXiv:2302.14816}, 2023.

\bibitem{shao2023nddepth}
Shuwei Shao, Zhongcai Pei, Weihai Chen, Xingming Wu, and Zhengguo Li.
\newblock {NDDepth}: Normal-distance assisted monocular depth estimation.
\newblock In {\em Proceedings of the IEEE/CVF International Conference on Computer Vision}, pages 7931--7940, 2023.

\bibitem{shao2024iebins}
Shuwei Shao, Zhongcai Pei, Xingming Wu, Zhong Liu, Weihai Chen, and Zhengguo Li.
\newblock {IEBins}: Iterative elastic bins for monocular depth estimation.
\newblock {\em Advances in Neural Information Processing Systems}, 36, 2024.

\bibitem{silberman2012indoor}
Nathan Silberman, Derek Hoiem, Pushmeet Kohli, and Rob Fergus.
\newblock Indoor segmentation and support inference from {RGBD} images.
\newblock In {\em Computer Vision--ECCV 2012: 12th European Conference on Computer Vision, Florence, Italy, October 7-13, 2012, Proceedings, Part V 12}, pages 746--760. Springer, 2012.

\bibitem{song2015sun}
Shuran Song, Samuel~P Lichtenberg, and Jianxiong Xiao.
\newblock {SUN RGB-D: A RGB-D} scene understanding benchmark suite.
\newblock In {\em Proceedings of the IEEE Conference on Computer Vision and Pattern Recognition}, pages 567--576, 2015.

\bibitem{teed2020raft}
Zachary Teed and Jia Deng.
\newblock {RAFT}: Recurrent all-pairs field transforms for optical flow.
\newblock In {\em Computer Vision--ECCV 2020: 16th European Conference, Glasgow, UK, August 23--28, 2020, Proceedings, Part II 16}, pages 402--419. Springer, 2020.

\bibitem{tian2024visual}
Keyu Tian, Yi Jiang, Zehuan Yuan, Bingyue Peng, and Liwei Wang.
\newblock Visual autoregressive modeling: Scalable image generation via next-scale prediction.
\newblock {\em arXiv preprint arXiv:2404.02905}, 2024.

\bibitem{van2017neural}
Aaron Van Den~Oord, Oriol Vinyals, et~al.
\newblock Neural discrete representation learning.
\newblock {\em Advances in Neural Information Processing Systems}, 30, 2017.

\bibitem{vaswani2017attention}
Ashish Vaswani, Noam Shazeer, Niki Parmar, Jakob Uszkoreit, Llion Jones, Aidan~N Gomez, {\L}ukasz Kaiser, and Illia Polosukhin.
\newblock Attention is all you need.
\newblock {\em Advances in Neural Information Processing Systems}, 30, 2017.

\bibitem{wang2023ord2seq}
Jinhong Wang, Yi Cheng, Jintai Chen, TingTing Chen, Danny Chen, and Jian Wu.
\newblock {Ord2Seq}: Regarding ordinal regression as label sequence prediction.
\newblock In {\em Proceedings of the IEEE/CVF International Conference on Computer Vision}, pages 5865--5875, 2023.

\bibitem{wang2024dcdepth}
Kun Wang, Zhiqiang Yan, Junkai Fan, Wanlu Zhu, Xiang Li, Jun Li, and Jian Yang.
\newblock {DCDepth}: Progressive monocular depth estimation in discrete {Cosine} domain.
\newblock {\em arXiv preprint arXiv:2410.14980}, 2024.

\bibitem{wang2019pseudo}
Yan Wang, Wei-Lun Chao, Divyansh Garg, Bharath Hariharan, Mark Campbell, and Kilian~Q Weinberger.
\newblock Pseudo-{LiDAR} from visual depth estimation: Bridging the gap in {3D} object detection for autonomous driving.
\newblock In {\em Proceedings of the IEEE/CVF Conference on Computer Vision and Pattern Recognition}, pages 8445--8453, 2019.

\bibitem{wei2023autoregressive}
Xing Wei, Yifan Bai, Yongchao Zheng, Dahu Shi, and Yihong Gong.
\newblock Autoregressive visual tracking.
\newblock In {\em Proceedings of the IEEE/CVF Conference on Computer Vision and Pattern Recognition}, pages 9697--9706, 2023.

\bibitem{wu2022nuwa}
Chenfei Wu, Jian Liang, Lei Ji, Fan Yang, Yuejian Fang, Daxin Jiang, and Nan Duan.
\newblock {N{\"U}WA}: Visual synthesis pre-training for neural visual world creation.
\newblock In {\em European Conference on Computer Vision}, pages 720--736. Springer, 2022.

\bibitem{yan2021videogpt}
Wilson Yan, Yunzhi Zhang, Pieter Abbeel, and Aravind Srinivas.
\newblock {VideoGPT}: Video generation using {VQ-VAE} and {Transformers}.
\newblock {\em arXiv preprint arXiv:2104.10157}, 2021.

\bibitem{yang2021transformer}
Guanglei Yang, Hao Tang, Mingli Ding, Nicu Sebe, and Elisa Ricci.
\newblock Transformer-based attention networks for continuous pixel-wise prediction.
\newblock In {\em Proceedings of the IEEE/CVF International Conference on Computer vision}, pages 16269--16279, 2021.

\bibitem{yang2024depth}
Lihe Yang, Bingyi Kang, Zilong Huang, Xiaogang Xu, Jiashi Feng, and Hengshuang Zhao.
\newblock Depth anything: Unleashing the power of large-scale unlabeled data.
\newblock In {\em Proceedings of the IEEE/CVF Conference on Computer Vision and Pattern Recognition}, pages 10371--10381, 2024.

\bibitem{yuan2022neural}
Weihao Yuan, Xiaodong Gu, Zuozhuo Dai, Siyu Zhu, and Ping Tan.
\newblock {NeW CRFs}: Neural window fully-connected {CRFs} for monocular depth estimation.
\newblock In {\em Proceedings of the IEEE/CVF Conference on Computer Vision and Pattern Recognition}, pages 3916--3925, 2022.

\bibitem{zeng2024wordepth}
Ziyao Zeng, Daniel Wang, Fengyu Yang, Hyoungseob Park, Stefano Soatto, Dong Lao, and Alex Wong.
\newblock {WorDepth}: Variational language prior for monocular depth estimation.
\newblock In {\em Proceedings of the IEEE/CVF Conference on Computer Vision and Pattern Recognition}, pages 9708--9719, 2024.

\bibitem{zhao2023unleashing}
Wenliang Zhao, Yongming Rao, Zuyan Liu, Benlin Liu, Jie Zhou, and Jiwen Lu.
\newblock Unleashing text-to-image diffusion models for visual perception.
\newblock In {\em Proceedings of the IEEE/CVF International Conference on Computer Vision}, pages 5729--5739, 2023.

\end{thebibliography}
}

\end{document}